%% file: main.tex
\pdfoutput=1
\documentclass[11pt]{article}

\usepackage[]{acl}

\usepackage{times}
\usepackage{latexsym}
\usepackage{colortbl}

\usepackage[T1]{fontenc}
\usepackage[utf8]{inputenc}

\usepackage{microtype}

\usepackage{inconsolata}

\usepackage{graphicx}

\usepackage{booktabs}
\usepackage[intlimits]{amsmath}  % place the subscripts and superscripts in the right position
\usepackage[colorinlistoftodos,size=tiny,disable]{todonotes}
\usepackage{caption}
\usepackage{pifont}
\usepackage{subcaption}
\usepackage{multirow}
\usepackage{enumitem}
\usepackage{makecell}
\usepackage{tabularx}
\usepackage{soul}
\usepackage{amssymb}

\newcommand{\hlc}[2][yellow]{{%
    \colorlet{foo}{#1}%
    \sethlcolor{foo}\hl{#2}}%
}

\newcommand{\se}[1]{\textcolor{black}{#1}}
\newcommand{\ycr}[1]{\textcolor{black}{#1}} % orange
\newcommand{\rz}[0]{\textit{E}}
\newcommand{\ro}[0]{\textit{N}}
\newcommand{\rth}[0]{\textit{G}$^{-}$\textit{(E)}}
\newcommand{\rtw}[0]{\textit{G}$^{+}$\textit{(N)}}

\newcommand{\bill}[0]{\texttt{Bill}$_{\texttt{en}}$}
\newcommand{\hansard}[0]{\texttt{Hansard}$_{\texttt{en}}$}
\newcommand{\deuparl}[0]{\texttt{DeuParl}$_{\texttt{de}}$}
\newcommand{\lynn}[0]{\texttt{EmoDefabel}$_{\texttt{de}}$}
\newcommand{\dagstuhl}[0]{\texttt{Dagstuhl}$_{\texttt{en}}$}

\title{Do Emotions Really Affect Argument Convincingness? \\ A Dynamic Approach with LLM-based Manipulation Checks}

\author{Yanran Chen and Steffen Eger \\
 NLLG, University of Technology Nuremberg (UTN) \\\url{https://nl2g.github.io/}\\
  \texttt{\{yanran.chen,steffen.eger\}@utn.de} \\
  }

\begin{document}

\maketitle

\begin{abstract}
Emotions have been shown to play a role in argument convincingness, yet this aspect is underexplored in the natural language processing (NLP) community. 
Unlike prior studies that use static analyses, focus on a single text domain or language, or treat emotion as just one of many factors, we introduce a dynamic framework inspired by manipulation checks commonly used in psychology and social science; leveraging LLM-based manipulation checks, this framework examines the extent to which perceived emotional intensity influences perceived convincingness.
Through human evaluation of arguments across different languages, text domains, and topics, we find that in over half of cases, human judgments of convincingness remain unchanged despite variations in perceived emotional intensity; when emotions do have an impact, they more often enhance rather than weaken convincingness.
We further analyze whether 11 LLMs behave like humans in the same scenario, finding that while LLMs generally mirror human patterns,
they struggle to capture nuanced emotional effects in individual judgments.
\end{abstract}

\input{structure/1_introduction}

\input{structure/2_related}

\input{structure/3_methods}

\input{structure/4_results}

\input{structure/6_conclusion}

\section*{Acknowledgements}

We thank the anonymous reviewers for their thoughtful feedback, which helped improve the paper. We also thank Jonas Belouadi, Daniil Larionov, Aida Kostikova, and Sotaro Takeshita for their valuable comments on the initial version of this paper, which greatly strengthened the work.
The NLLG Lab gratefully acknowledges support from the Federal Ministry of Education and Research (BMBF) via the research grant
“Metrics4NLG” and the German Research Foundation (DFG) via the Heisenberg Grant EG 375/5-1. This project has been conducted as part of the EMCONA (The Interplay of Emotions and Convincingness in Arguments) project, which is funded
by the DFG (project EG-375/9-1). We also thank Roman Klinger and Lynn Greschner for their valuable feedback in the context of the EMCONA grant, which supports our collaborative research papers.

\bibliography{custom}
%\bibliographystyle{acl_natbib}

%\onecolumn

\appendix
\input{structure/999_appendix}

\end{document}

%% file: structure/1_introduction.tex
\section{Introduction}

\begin{figure}[!t]
    \centering
    \includegraphics[width=\linewidth]{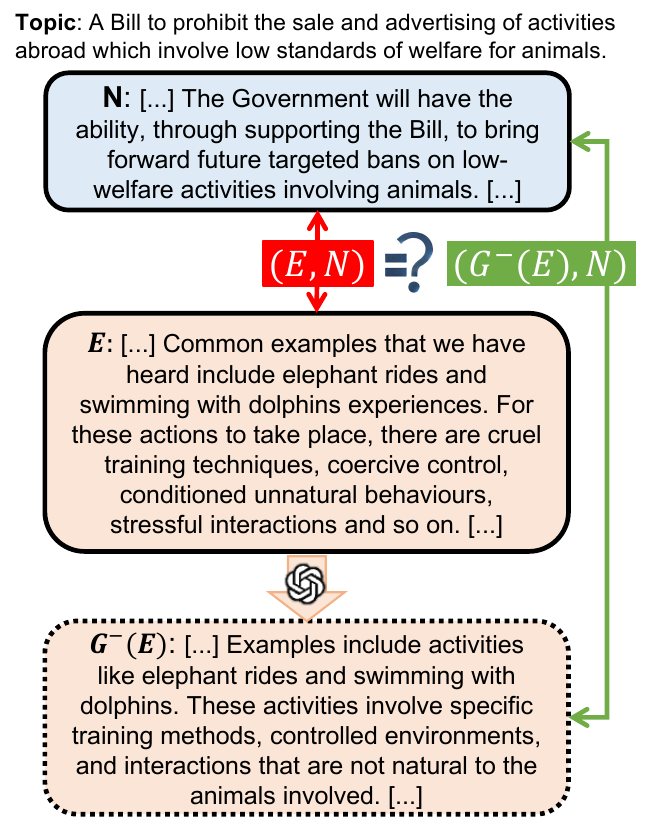}
    \caption{An example test case. \protect\rz{} is an argument with emotions and \protect\ro{} is an argument without emotions, both addressing the same topic with the same stance. \protect\rth{} is a counterpart of \protect\rz{} with reduced emotion. We compare the convincingness ranking of the pair (\protect\rz{}, \protect\ro{}) to that of the pair (\protect\rth{}, \protect\ro{}) to observe the effect of emotions on argument convincingness in a dynamic way.}
    \label{fig:example}
    \vspace{-.6cm}
\end{figure}

Emotional appeals have long been recognized as a core component of persuasion \citep{konat2024pathos,habernal2017argumentation}.
Aristotle’s triad of logos, ethos, and pathos \citep{kennedy1991theory} 
emphasizes the multifaceted nature of effective rhetoric. While logical reasoning (\emph{logos}) and the speaker's credibility (\emph{ethos}) are essential, the ability to evoke emotions in the audience (\emph{pathos}) 
\se{may also be crucial} 
\se{in order to} 
make the audience more receptive to the arguments \citep{wachsmuth-etal-2017-computational}.

Despite active research on 
argumentation and argument quality 
in the NLP community \citep[e.g.][]{habernal-gurevych-2016-makes,habernal-gurevych-2016-argument,gleize-etal-2019-convinced,wan-etal-2024-evidence,rescala-etal-2024-language,eger-etal-2017-neural,wachsmuth-etal-2017-computational,wachsmuth2024argument},
the 
pathos dimension has received 
\se{undeservedly little}
attention 
 \citep{evgrafova-etal-2024-analysing,greschner2024fearfulfalconsangryllamas}; emotional appeal is often discussed as a logical fallacy in arguments \citep[e.g.,][]{vijayaraghavan-vosoughi-2022-tweetspin,goffredo-etal-2023-argument,li-etal-2024-reason,mouchel2024logicalfallacyinformedframeworkargument}.
Existing NLP studies exploring the interplay between emotions and \emph{\se{argument} convincingness} often lack a specific focus on the emotional dimension and fail to control for confounding factors \citep[e.g.][]{habernal-gurevych-2016-argument,habernal2017argumentation,wachsmuth-etal-2017-computational}. A confounder refers to a variable that influences both the independent variable (the factor being manipulated: emotions) and the dependent variable (the outcome being measured: convincingness), potentially distorting the observed relationship between them.
\ycr{
To address this gap, we propose a \emph{dynamic} approach that systematically varies emotional intensity to observe its effect on argument convincingness, following the logic of psychological manipulation checks \citep{hoewe2017manipulation,ejelov2020rarely}. Here, emotional intensity is the manipulated variable and convincingness the dependent one. We call it dynamic because it captures the effect of varying emotional intensity on convincingness, moving beyond static comparisons between fixed argument pairs.}
To achieve this, we leverage LLMs to rephrase an argument to generate a counterpart that %is more/less likely to 
evokes stronger/weaker emotions, and then compare its convincingness to the original argument, thereby minimizing the effect of confounders. The judgments are evaluated relative to an anchor argument \ycr{(\S\ref{sec:human})}, as illustrated in Figure \ref{fig:example}, to obtain more reliable subjective human evaluation  
\citep{zhang2017improved,gienapp-etal-2020-efficient,jin-etal-2022-logical,habernal-gurevych-2016-argument}. 
This framework enables us to examine how variations in perceived emotional intensity influence judgments of convincingness for a given argument in a controlled manner. 
\ycr{To test the robustness of our findings across annotation setups, we further use Likert-scale ratings of argument convincingness as a simpler alternative to the relative, anchor-based evaluation (\S\ref{sec:robust}). Although absolute ratings are generally considered less reliable for such subjective evaluation, our observations remain consistent, demonstrating the robustness of both our dynamic approach and findings.}

Besides, we move beyond prior studies that focus predominantly on English arguments or single-domain datasets \citep{habernal-gurevych-2016-argument,habernal2017argumentation,wachsmuth-etal-2017-computational,greschner2024fearfulfalconsangryllamas}. We expand the scope to explore both English and German arguments across diverse text domains, including political debates, online portals, and curated human-written arguments. Our multilingual and cross-domain analysis provides a comprehensive view of how perceived emotional intensity 
affects convincingness across different contexts.

Finally, inspired by recent studies exploring cognitive biases in LLMs \citep{10.1093/pnasnexus/pgae233,echterhoff-etal-2024-cognitive,10.1162/tacl_a_00673,macmillan2024ir}, we further investigate whether LLMs behave like humans when judging argument convincingness under the influence of emotional `bias'. Although emotion is not always considered a fallacy or bias in argumentation \citep{walton2005fundamentals,Duckett02042020,evgrafova-etal-2024-analysing}, understanding its impact on argument evaluation is crucial for developing 
models intended for automated argument evaluation \citep[e.g.,][]{wachsmuth2024argument,rescala-etal-2024-language,mirzakhmedova2024large}.
\noindent\textbf{Our contributions are:}\footnote{Code+data: \url{https://github.com/cyr19/argument_emotion_llm_manipulation}}  
\begin{itemize}[topsep=2pt,itemsep=-1pt,leftmargin=*]
    \item We propose a novel framework to analyze how emotions influence perceived convincingness in a controlled manner. Our findings show that in over half of cases, human judgments remain unaffected by emotional intensity, while emotions more often enhance rather than weaken convincingness.
    
    \item We demonstrate that LLMs can effectively modify the emotional impact of arguments while preserving their original meaning, enabling precise comparisons of argument emotions.
    
    \item We conduct a multilingual, cross-domain analysis, showing that (i) when topics and domains align, emotions impact convincingness similarly in German and English, and (ii) emotions are more likely to enhance convincingness in political debates than in other domains. 
    
    \item We investigate whether LLMs exhibit human-like preferences in evaluating argument convincingness, particularly regarding emotions. While they broadly mirror human patterns, they fail to capture nuanced emotional effects in individual judgments.
\end{itemize}

%% file: structure/2_related.tex
\section{Related Work}
This work primarily connects to (1) the interplay between emotions and argument convincingness, while also relating to (2) human-like biases in LLMs.

\paragraph{Emotion vs.\ convincingness}
Emotions have been shown to play a role in argument convincingness in both fields of computational argumentation \citep[e.g.][]{habernal-gurevych-2016-argument,wachsmuth-etal-2017-computational,greschner2024fearfulfalconsangryllamas} and philosophy/psychology \citep[e.g.][]{kennedy1991theory,konat2024pathos,benlamine2015emotions}. 

In NLP, emotional appeal is primarily studied within the context of logical fallacy in arguments \citep{evgrafova-etal-2024-analysing} or as a secondary focus in relation to argument convincingness \citep{greschner2024fearfulfalconsangryllamas}. 
The most relevant works include: \citet{habernal-gurevych-2016-argument} find that human annotators identify emotional aspects as positively contributing to argument convincingness. \citet{habernal2017argumentation} introduce an emotional appeal layer in a modified Toulmin argumentation model, showing that 6\% of arguments are purely emotional. \citet{wachsmuth-etal-2017-computational} analyze arguments across 15 dimensions, finding a weak positive correlation between emotional appeal and convincingness. \citet{lukin-etal-2017-argument} demonstrate that audience-specific factors improve belief change prediction, particularly for emotional arguments. \citet{greschner2024fearfulfalconsangryllamas} examine specific emotions, \ycr{showing that arguments expressing joy and pride are rated as more convincing, while those expressing anger are rated as less convincing.}

Previous studies rely on fixed analyses that do not control for confounders. 
In contrast, we adopt a dynamic approach, controlling for confounding factors and examining how perceived convincingness changes with varying emotional intensity. Our methodology aligns with psychological manipulation checks \citep{hoewe2017manipulation,ejelov2020rarely}, treating emotional intensity as the manipulated variable and convincingness as the dependent variable.

Additionally, prior work has largely focused on English, except for \citet{greschner2024fearfulfalconsangryllamas}, who examine German arguments. Since emotional effects may vary across cultures, we study both English and German arguments. We also expand the scope by incorporating diverse text domains, including political debates, online portals, and curated human-written arguments, unlike previous studies limited to a single domain.

%\vspace{-.2cm}
\paragraph{Human-like biases in LLMs}
An array of studies has demonstrated human-like biases in LLMs \citep[e.g.,][]{liang2021towards,echterhoff-etal-2024-cognitive,10.1162/tacl_a_00673}. Social biases, such as sentiment, stereotype, and gender biases, have been extensively investigated \citep[e.g.,][]{huang-etal-2020-reducing,nadeem-etal-2021-stereoset,10.1145/3582269.3615599,viswanath2023fairpytoolkitevaluationsocial}.

Beyond social biases, LLMs also mimic human cognitive biases in reasoning and decision-making \citep{10.1093/pnasnexus/pgae233,hagendorff2023human,talboy2023challengingappearancemachineintelligence,echterhoff-etal-2024-cognitive,10.1162/tacl_a_00673,sumita2024cognitivebiaseslargelanguage,macmillan2024ir}. For instance, \citet{10.1093/pnasnexus/pgae233} show that LLMs, like humans, perform better when task semantics align with logical inference (`content effect’). Similarly, \citet{echterhoff-etal-2024-cognitive} find LLMs exhibit decision-making biases such as anchoring bias \citep{tversky1974judgment}, status quo bias \citep{samuelson1988status}, and framing bias \citep{tversky1974judgment}. Meanwhile, \citet{macmillan2024ir} analyze LLMs’ irrationality across 12 cognitive tasks \citep{kahneman1972subjective,bruckmaier2021tversky}, revealing both human-like errors and distinct deviations.

Although emotional appeal is not inherently a bias or fallacy but a persuasion strategy, it is crucial to examine whether LLMs' preferences align with human judgments, especially given their growing role in argument evaluation \citep[e.g.,][]{wachsmuth2024argument,rescala-etal-2024-language,mirzakhmedova2024large}. Inspired by studies on cognitive biases in LLMs, we investigate whether LLMs exhibit human-like behavior in how emotional intensity influences argument convincingness.

%% file: structure/3_methods.tex
%\vspace{-.2cm}
\section{Evaluation Setup}\label{sec:setup}
%\vspace{-.2cm}

We employ a dynamic framework to explore how the intensity of emotions evoked in readers impacts their judgments of convincingness. 
In this work, \textbf{we treat emotional intensity as the overall strength of emotions felt by readers, without considering specific emotions}.
We follow previous works 
\citep{habernal-gurevych-2016-argument,toledo-etal-2019-automatic} to leverage pairwise comparisons for evaluation \se{because }it yields more reliable annotations compared to the absolute ratings, especially for such subjective evaluation tasks 
\citep{zhang2017improved,10.1162/coli_a_00426,gienapp-etal-2020-efficient}. 

Our setting is as Figure \ref{fig:example} shows:  
among one pair of arguments
that share \emph{the same stance} on a given topic but \emph{differ in their content}, 
\textbf{\rz{}}  
is \se{(set up to be)} 
\textbf{emotion-evoking}, while 
\textbf{\ro{} 
\se{does not (typically)} evoke emotions}.
We then use LLMs to generate a counterpart argument for $E$, 
\textbf{\rth{}}, 
which retains the same meaning as \rz{} but \textbf{evokes less emotion}. To inspect how perceived argument convincingness is affected by emotions, we compare the convincingness ranking of (\rth{},\ro{}) to that of the original pair (\rz{},\ro{}). 
The reason to not compare the arguments with a similar content, i.e., \rz{} vs.\ \rth{} is that we want 
to minimize the effect of human's prior belief about whether emotions should contribute to argument convincingness.
Analogously, we generate a counterpart for \ro{}, \textbf{\rtw{}}, with \textbf{increased emotional intensity} and observe how the convincingness ranking changes from (\rz{}, \ro{}) to (\rz{}, \rtw{}). Finally, we include the fully LLM-generated pair (\rth{}, \rtw{}) in our evaluation. 

The goal \se{of $G^+/G^-$} is to maintain the core meaning of the argument while modifying its emotional appeal. Although humans could be used to create such counterparts \citep[e.g.][]{huffaker2020crowdsourced,velutharambath_wuehrl_klinger_2024a}, 
this approach is largely impractical at scale because it is costly.  Instead, we use LLMs to efficiently generate required variations and assess their \ycr{ability to perform this task through human evaluation.}

Thus, \textbf{for each original argument pair \se{$(E,N)$}, we create three counterpart pairs} with varying levels of emotional intensity, resulting in a total of four argument pairs per test instance. The original argument pair serves as the anchor, from which we see how the convincingness rankings of the other argument pairs change.
We list 
all possible change scenarios in Table \ref{tab:comparison} 
and 
divide 
them into three categories: 
\ref{tab:comparison}: 
%\begin{itemize}
    %\item 
    (1) \textbf{Consistent}: convincingness ranking \emph{does not change} with varying emotional intensities. 
    (2) \textbf{Positive}: 
    an argument is perceived as more/less convincing when it evokes stronger/weaker emotions 
    \se{ 
    (convincingness and emotionality have the same directionality).}
    %\item 
    (3) \textbf{Negative}: 
    an argument is perceived as more/less
    convincing when it evokes weaker/stronger emotions,
    and less convincing when it evokes stronger emotions
    \se{%
    (convincingness and emotionality have the opposite directionality).}
%\end{itemize}

\begin{table}[!t]
    \centering
    \resizebox{.75\linewidth}{!}{
    \begin{tabular}{@{}cc>{\columncolor[RGB]{153,255,153}}c|>{\columncolor[RGB]{255,204,204}}cc>{\columncolor[RGB]{153,255,153}}c|>{\columncolor[RGB]{255,204,204}}cc@{}}
         \toprule
         Argument Pair & \multicolumn{7}{c}{Convincingness Ranking} \\ \midrule
Anchor: (\rz{}, \ro{})   & \multicolumn{2}{c|}{>} & \multicolumn{3}{c|}{=}            & \multicolumn{2}{c}{<}    \\ \midrule
(\rth{}, \ro{})  & > & $\le$ & > & = & < & $\ge$ & < \\
(\rz{}, \rtw{})  & > & $\le$ & > & = & < & $\ge$ & < \\
(\rth{}, \rtw{}) & > & $\le$ & > & = & < & $\ge$ & < \\ \bottomrule
\end{tabular}}
    \iffalse
    \includegraphics[width=\linewidth]{structure/figs/comparison.pdf}
    \fi
    \caption{All convincingness change scenarios. Cells marked in green indicate positive cases, red indicates negative cases, and consistent cases are left with a white background. \se{Math relation symbols $>,<,=$ refer to convincingness.}}
    \label{tab:comparison}
    \vspace{-.5cm}
\end{table}

The first 
row in the table presents all possible convincingness ranking\se{s} of the original argument pair (\rz{},\ro{}). The subsequent rows show the convincingness rankings of the counterpart argument pairs where the emotional intensity of the argument on the left has been reduced (\rth{},\ro{}), that of the argument on the right has been increased (\rz{},\rtw{}), 
or both (\rth{},\rtw{}). Cells highlighted in \hlc[green!50]{green} indicate cases where the convincingness of the left argument decreases as its emotional intensity decreases \emph{relative} to the right argument, 
suggesting a \hlc[green!50]{\textbf{positive}} impact of emotions on convincingness. 
This occurs when the convincingness ranking shifts from the left being $>$ to $\leq$ the right argument, or from being $=$ to $<$ the right argument. 
Conversely, cells highlighted in \hlc[pink]{red} indicate cases where the convincingness of the left argument increases as its emotional intensity decreases \emph{relative} to the right argument, reflecting a \hlc[pink]{\textbf{negative}} impact of emotions on convincingness.
Finally, cases where the convincingness rankings remain \hlc[white]{\textbf{consistent}} 
\se{retain} 
a \hlc[white]{white} background.

\vspace{-.1cm}
\paragraph{Metrics}
For each instance \se{$(E,N)$}, we calculate the percentages of consistent, positive, and negative cases. We then average the percentages of each category across all test instances to derive three metrics that indicate the overall frequencies of the three categories in humans. 
We call the metrics: \textbf{consistency rate}, \textbf{positivity rate}, and \textbf{negativity rate}. Their formulas are as follows:

\vspace{-.2cm}
\begin{align}
    Rate_{category} = \frac{1}{n} \sum_{i=1}^{n} \frac{C_{category, i}}{3},
\end{align}
where $n$ is the total number of test instances, $\textit{C}_{\textit{category}}$ is the count  
of cases in the specified category for the $i$-th instance, and $category \in \{\text{consistent, positive, negative}\}$.

\section{Dataset Construction}

We source \textbf{50 anchor argument pairs} from each of five datasets (\S\ref{sec:source}) and utilize GPT4o\footnote{\url{https://openai.com/index/gpt-4o-system-card/}} to generate their \textbf{counterparts} with variations in emotional intensity (\S\ref{sec:synthetic}).
%\vspace{-.2cm}
\subsection{Anchor: \rz{} \& \ro{}}\label{sec:source}
We leverage two established datasets which have human annotations for argument convincingness and emotions, \dagstuhl{} \citep{wachsmuth-etal-2017-computational} and \lynn{} \citep{greschner2024fearfulfalconsangryllamas}. 
Besides, we create three datasets ourselves from political debates, \bill{}, \hansard{}, and \deuparl{}, since emotional appeal is a common strategy used by politicians to influence perceptions and decisions \citep{brader2005striking}; this domain is therefore expected to be rich in emotional content. 
From each data source, we select 50 argument pairs where \textbf{\rz{} is more likely and \ro{} is less likely to evoke emotions}. The subscripts in the dataset names indicate the language: `en' for English and `de' for German. \se{In the following, we }describe how we extract argument pairs from each data source.

\subsubsection{Arguments from Political Debates}\label{sec:data_ours} 

We \textbf{crawl} parliamentary debates for \hansard{} from the UK Hansard\footnote{\url{https://hansard.parliament.uk/}} and for \deuparl{}\footnote{We name it DeuParl following previous studies leveraging this corpus \citep[e.g.][]{walter2021diachronicanalysisgermanparliamentary,kostikova-etal-2024-fine,chen2024syntacticlanguagechangeenglish}.} from 
the German Bundestagsprotokolle.\footnote{\url{https://www.bundestag.de/protokolle}}
The datasets cover the past 3–5 years.\footnote{Hansard: 2022/01/05-2024/07/19; German Bundestagsprotokolle: 2020/01/15-2024/09/27}
We heuristically \textbf{segment} each speech into balanced-length paragraphs. The original crawled texts are divided by double line breaks. If a paragraph has fewer than 60 tokens or the next one has fewer than 20 tokens or starts with a left bracket, we merge them cumulatively. From these processed paragraphs, we select argumentative texts for evaluation.

In our \textbf{pilot annotations} with \hansard{}, we find that within a single debate on a broad topic, diverse subtopics make it difficult to pair arguments with the same topic. Additionally, the interactive nature of debates complicates determining a paragraph’s focus without context.
To address this, we first conduct \textbf{pre-annotation} on a small scale for five \emph{Second Reading debates of Bills} relevant to family and animals,\footnote{\url{https://www.parliament.uk/about/how/laws/passage-bill/commons/coms-commons-second-reading/}} which are easier to annotate because the Bill debated provides a clear topic. We then refine GPT4o prompts to develop \textbf{classifiers} for identifying argument pairs that share a topic and stance but differ in emotional appeal. The final classifiers achieve precisions of 0.80 (English) and 0.76 (German) for detecting topic-aligned arguments and a macro F1 of $\sim$0.75 for distinguishing emotional from non-emotional arguments.
See Appendix \ref{app:pre} for details.

%\vspace{-.3cm}
\paragraph{\bill{}}
From the argument pairs labeled as having the same topic and stance during the pre-annotation phase, we randomly sample 50 pairs, with one argument labeled as emotion-evoking and the other as non-emotion-evoking. The topic for each argument pair is the brief introduction of the Bill crawled.

%\vspace{-.3cm}
\paragraph{\hansard{} \& \deuparl{}}
Debates are filtered using pre-selected keywords related to recent wars, refugee crises, and migration (see Table \ref{tab:keywords} in the appendix for the full list), as these highly debated topics are likely to evoke strong emotions. 
For \hansard{}, we retain debates whose titles contain these keywords. For \deuparl{}, we include debates whose introductions mention the keywords.
Finally, an annotator from the pre-annotation phase selects 50 argument pairs from the candidates  
\ycr{selected out by the GPT4o classifiers}
for both \hansard{} and \deuparl{}. These argument pairs are \textbf{manually verified} to meet our criteria --- both arguments address the same topic with the same stance but differ in their emotional aspect. 
A human-written topic is assigned to each pair.

%\vspace{-.2cm}
\subsubsection{Arguments from others}
%\vspace{-.1cm}
We randomly select 50 argument pairs from each of \textbf{\dagstuhl{}} and \textbf{\lynn{}} that meet our criteria, based on the emotion annotations in the original works. See Appendix \ref{app:other} for details.

%\vspace{-.1cm}
\subsection{Counterpart: \rth{} \& \rtw{}}\label{sec:synthetic}
%\vspace{-.1cm}
 
We leverage \textbf{GPT4o}\footnote{We used the version `gpt-4o-2024-08-06' with a temperature of 0.6 and a top\_p of 0.9 for GPT4o. The randomness was set to a moderate level to balance creativity and consistency, as the task involves generating content similar to creative writing while ensuring the meaning of the original argument is preserved.} to synthesize our counterpart arguments, namely \textbf{\rth{}} and \textbf{\rtw{}}. Specifically,
we prompt GPT4o (zero-shot) to either introduce or remove emotions by \textbf{rephrasing} the original arguments, using the prompts listed in Table \ref{tab:prompt_synthetic} (appendix),
since we aim for counterpart arguments that convey the same information as the original ones. During generation, if the output does not receive the expected label from the \ycr{binary} emotion classifiers used in \S\ref{sec:data_ours}, the process is repeated for up to five rounds. 

We randomly sample five argument pairs (original + synthetic) for each direction (introducing or removing emotions) from each dataset, totaling 50 argument pairs for content preservation \textbf{evaluation}. Each pair is rated by three crowdworkers for content similarity on a Likert scale of 1–5. The pairs receive an average score of 4.5, where 4 denotes ‘Same Claims, Minor Content Differences’ (minor details differ, but no major evidence changes), and 5 represents ‘Identical Content, Different Style/Tone’ (only rhetorical or emotional differences). Thus, we conclude that the main message is well preserved throughout the process.
The effectiveness of adjusting emotional appeal is further evaluated in our primary human study (\S\ref{sec:human}).

\begin{table*}[t]
\setlength\tabcolsep{2pt} 
\resizebox{\linewidth}{!}{
\begin{tabular}{@{}lcccc|cc|ll@{}}
\toprule
\textbf{Dataset} & \textbf{Lang} & \textbf{\#Instances} & \textbf{\#Pairs} & \textbf{\#Arguments} & \textbf{\#Tokens} & \textbf{\#Sents} & \textbf{Domain}              & \textbf{Topics}                   \\ \midrule
\bill{}    & en       & 50           & 200      & 128          & 147.4     & 6.1      & Parliamentary debates   & Bills related to family and animals                \\
\hansard{}          & en       & 50           & 200      & 154          & 159.3     & 6.4      & Parliamentary debates        & Refugees, wars, migrants          \\
\dagstuhl{}         & en       & 50           & 200      & 128          & 86.8      & 4.5      & Online portal                & -                                 \\
\deuparl{}         & de       & 50           & 200      & 126          & 144.3     & 7.4      & Parliamentary debates        & Refugees, wars, migrants          \\
\lynn{}          & de       & 50           & 200      & 160          & 92.8      & 4.5      & Curated human-written arguments & Health, law, finance and politics \\ \midrule
Total/Average  & -        & 250          & 1,000     & 696          & 126.1     & 5.7      & -                            & -                                 \\ \bottomrule
\end{tabular}}
\caption{Metadata of datasets used in this work. \textbf{Left}: number of test instances, argument pairs, and unique arguments. \textbf{Middle}: average number of tokens and sentences per argument, measured with the Stanza tokenzier \citep{qi-etal-2020-stanza}. \textbf{Right}: domains and topics of the datasets.}\label{tab:data}
%\vspace{-.6cm}
\end{table*}

\subsection{Final Datasets}
Our final datasets comprise 250 test instances, each consisting of one original argument pair and three counterpart pairs. The datasets include both English and German texts, spanning various domains and topics. The metadata of the datasets is summarized in Table \ref{tab:data}. 

%\vspace{-.2cm}
\section{Human Annotation}
%\vspace{-.2cm}
%\subsection{Annotation}

We randomly divide the 50 instances (200 argument pairs) from each dataset into 10 batches, each with 5 instances (20 argument pairs). 
\ycr{Every batch is annotated by 5 individuals. }
One annotator evaluates at least one batch, 
allowing us to calculate inter-annotator agreements and base observations on individual annotators. 
 Although our primary focus is on how convincingness rankings change, we also include comparisons of emotional intensity to evaluate whether GPT4o adjusts the emotional appeal of arguments as intended.

Annotators compare emotions and convincingness of one argument pair by answering two \textbf{subjective} questions:
%\begin{itemize}
    %\item 
    (i) \textbf{Convincingness}: \emph{Which argumentative text \se{do} you find more convincing?}
    %\item 
    (ii) \textbf{Emotion}: \emph{Which argumentative text evokes stronger emotions in you?}
%\end{itemize}
\textbf{Equivocal judgments} are allowed, i.e., 
annotators 
can judge both arguments as equally convincing or evoking an equal level of emotion.
During annotation, argument pairs are shown with their topics. See Appendix \ref{app:anno} for screenshots of the annotation interface.

%\vspace{-.2cm}
\paragraph{Annotators}
We hire annotators from two sources: university students  
\se{and}
the crowd-sourcing platform Prolific:\footnote{\url{https://www.prolific.com/}}
\begin{itemize}[topsep=2pt,itemsep=-1pt,leftmargin=*]
    \item \textbf{Student}: 
4 students 
are hired for this task. All annotators possess fluent to native-level proficiency in the languages of the evaluated arguments and are all based in Germany. One of them is a PhD student, and the others are Master's students. Three of them are involved 
in the pre-annotation phase to select out the needed argument pairs.
    \item \textbf{Crowdsourcing}: 
As our dataset includes arguments from political debates, we assume native speakers in the corresponding countries provide more reliable annotations. Thus, we use Prolific's \textbf{prescreening} to select native English/German speakers in the UK/Germany. Furthermore, to filter out individuals who may randomly fill in their profiles, participants are asked to re-rate their language proficiency, 
and those with inconsistent responses are \textbf{screened out} from the tasks. We also include \textbf{three attention checks} by randomly inserting instruction sentences, such as `select the answer whose first number equals three minus two', into the arguments. 
Overall, 38\% of the crowdworkers fail at least two attention checks, and their submissions are excluded from our analysis. 
This process is repeated iteratively until we obtain sufficient submissions for each batch.
\end{itemize}
\vspace{.2cm}
We summarize the number of student and crowdsourcing annotators for each dataset in Table \ref{tab:annotator} (left side)%
; the values indicate the total annotators involved in annotating each batch. The total annotation cost is around 1,500 Euros.

\begin{table}[]
\vspace{-.2cm}
\centering
\setlength\tabcolsep{2pt} 
\resizebox{\linewidth}{!}{%
\begin{tabular}{@{}lcc|cccccc@{}}
\toprule
& \multicolumn{2}{c|}{\textbf{\#Annotators}} & \multicolumn{6}{c}{\textbf{Agreements}} \\
              & \textbf{S} &  \textbf{C} & \multicolumn{3}{c}{\textbf{EMO}} & \multicolumn{3}{c}{\textbf{CONV}} \\ %\midrule
              &&& $\alpha$ & Full & Maj. & $\alpha$ & Full & Maj. \\ \midrule
\dagstuhl{}      & 1        & 4         &  0.506  & 6.5\% & 74.5\%   & 0.540  & 14.0\% & 80.0\%   \\
\bill{} & 1        & 4        &  0.449  & 7.0\% & 76.5\%   & 0.463  & 10.5\% & 78.0\%   \\
\hansard{}       & 1        & 4        &  0.361 & 0.5\% & 68.0\%   & 0.371  & 6.0\% & 75.0\%   \\
\lynn{}       & 2        & 3       &  0.729 & 13.5\% & 87.5\%   & 0.607  & 16.0\% & 82.0\%   \\
\deuparl{}       & 3        & 2       & 0.352  & 8.0\% & 80.5\%   & 0.364  & 4.5\% & 74.5\%   \\ \midrule
Avg    & -        & - & 0.479 & 7.1\% & 77.4\% & 0.469 & 10.2\% & 77.9\% \\ 
\bottomrule
\end{tabular}}
\caption{\textbf{Left}: Number of student (S) and crowdsourcing (C) annotators per batch. \textbf{Right}: Krippendorf's $\alpha$ for the most agreeing annotator pairs (\textbf{$\alpha$}), the percentages of annotation instances where all annotators agree on a certain label (\textbf{Full}),  and the percentage of annotation instances where at least three annotators agree on a certain label (\textbf{Maj.}). 
}
\label{tab:annotator}
%\vspace{-.6cm}
\end{table}

%\vspace{-.3cm}
\paragraph{Inter-annotator agreement}
\se{While} we acknowledge the inherent subjectivity in evaluating emotion and convincingness, 
we report inter-annotator agreement to present the level of consistency in these evaluations of emotional intensity (EMO) and convincingness (CONV).
Following \citet{wachsmuth-etal-2017-computational}, in Table \ref{tab:annotator} (right)% in the appendix
, we report the Krippendorf's $\alpha$ agreement \citep{castro-2017-fast-krippendorff} for the most agreeing annotator pairs\footnote{We average the agreements of the most agreeing annotators over batches per dataset since our sample size for calculating agreements is much smaller than \citet{wachsmuth-etal-2017-computational} (20 vs.\ 320).} (column `$\alpha$'), the percentages of annotation instances where all annotators agree on a certain label (column `Full'), and the percentages of annotation instances yielding a valid majority vote (column `Maj.'). 
The agreement among the most agreeing annotator pairs ranges from 0.352 to 0.729 for EMO and from 0.364 to 0.607 for CONV.
Full agreements are only up to 16.0\%, while 
majority agreements range from moderate to high across different datasets, with 68\% to 87.5\% for EMO and 62\% to 85\% for CONV. This suggests a decent level of annotation agreement, considering that \citet{wachsmuth-etal-2017-computational} reported 94.4\% majority agreement and a Krippendorff's $\alpha$ of 0.26–0.45 for the most agreeing annotator pairs when evaluating emotional appeal and argument effectiveness on a Likert scale of 1–3; both tasks can also be seen as three-way classifications similar to ours but involved only three annotators. 
\ycr{However, we note that when computing agreement in a more standard way --- by averaging across all annotators and batches --- the agreement decreases to around 0.2 Krippendorff’s $\alpha$ for both criteria. To validate the robustness of our findings, we conducted an additional annotation study on a small subset of data using a completely different setup and explore whether we can draw similar conclusions in \S\ref{sec:robust}.}

%% file: structure/4_results.tex
\begin{figure}[!t]
    \centering
    \includegraphics[width=1.07\linewidth]{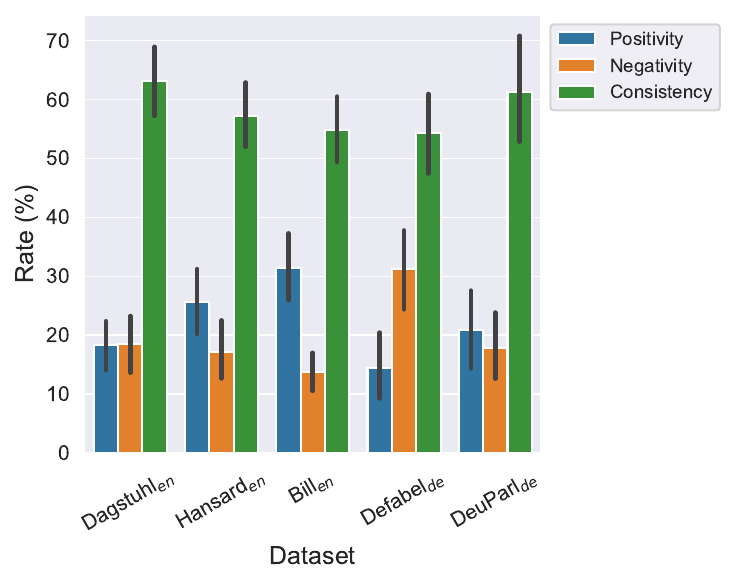}
    \vspace{-1cm}
        \caption{Consistency, positivity, and negativity rates of human judgments on convincingness. 
        }
    \label{fig:percent}
    \vspace{-.5cm}
\end{figure}

\subsection{Evaluation Results}\label{sec:human}

\paragraph{Effectiveness of GPT4o in adjusting emotional appeal}
We evaluate whether \rth{} evokes weaker emotion than \rz{} and whether \rtw{} evokes stronger emotion than \ro{}, as intended. To do so, we compute best-worst scaling (BWS) scores for each of the four argument groups based on emotion comparison annotations. Majority votes from the five annotators are used; if none exists, equivalent judgments are considered. While arguments with similar content (e.g., \rz{} vs.\ \rth{}) are not directly compared with each other, both are evaluated against the other two arguments within the same instance, making the BWS-based comparison between the arguments within the same content meaningful. Higher scores reflect greater perceived emotional intensity. Table \ref{tab:bws} (appendix) 
presents the BWS scores, showing that \rz{} consistently scores higher than \rth{} and \ro{} lower than \rtw{} across datasets. This suggests that \emph{GPT4o is overall effective in modifying arguments to be more or less emotion-evoking as intended}, supporting the premise for analyzing changes in convincingness rankings. 

%\vspace{-.2cm}
\paragraph{Do emotions really affect convincingness?} 
Figure \ref{fig:percent} illustrates the consistency, positivity, and negativity rates. We present the averages across individual annotators, with error bars representing 95\% confidence intervals. 
While the metrics vary across datasets and domains, we observe that consistency achieves the highest rates consistently across datasets, roughly ranging from 54\% to 62\%. This indicates that, \textbf{in more than half of the cases, humans are not influenced by variations in perceived emotions when judging convincingness}. In political debate domain datasets --- \hansard{}, \bill{}, and \deuparl{} --- positive rates are consistently higher than negative rates, averaging an 8-percentage-point difference.
In contrast, in \dagstuhl{}, positive and negative rates are roughly equal ($\sim$18\%), whereas in \lynn{}, negative rates dramatically exceed positive rates (30\% vs.\ 14\%). These differences may be attributed to variations in dataset domains and argument topics. In Appendix \ref{app:anno}, we show examples where emotions have positive/negative impacts on argument convincingness from \hansard{}/\lynn{}; in \lynn{}, topics often require more factual evidence, making emotions less influential or even detrimental.
Finally, we observe slight differences between the English and German datasets, using \hansard{} and \deuparl{} as examples, where argument topics are similar and both originate from political debates: the rates are overall comparable, with German being less affected by emotions (consistency rates: 60\% vs.\ 56\%) and also less positively influenced by emotions (positivity rates: 20\% vs.\ 25\%) compared to English.

\begin{table}[!t]
\centering
\setlength\tabcolsep{1.5pt} 
\resizebox{.9\linewidth}{!}{
\begin{tabular}{@{}llc@{}}
\toprule
\textbf{Model Family} & \textbf{Checkpoint}                           & \textbf{Size} \\ \midrule
OpenAI%\footnote{{\url{https://platform.openai.com/docs/models}}}      
& gpt-3.5-turbo                        & -    \\
\citep{openai2024gpt4technicalreport}             & gpt-4o-mini                          & -    \\
             & gpt-4o-2024-08-06                    & -    \\ \midrule
Llama3%\footnote{\url{https://huggingface.co/meta-llama}}
& Llama-3.2-1B-Instruct     & 1B   \\
\citep{grattafiori2024llama3herdmodels}             & Llama-3.2-3B-Instruct     & 3B   \\
             & Llama-3.3-70B-Instruct    & 70B  \\ \midrule
Qwen2.5%\footnote{\url{https://huggingface.co/Qwen}} 
& Qwen2.5-0.5B                    & 0.5B \\
\citep{qwen2.5}             & Qwen2.5-7B-Instruct             & 7B   \\
           & Qwen2.5-72B-Instruct            & 72B  \\ \midrule
Mistral%\footnote{\url{https://huggingface.co/mistralai}}
& &  \\
\citep{jiang2023mistral7b} & Mistral-7B-Instruct-v0.3   & 7B   \\
\citep{jiang2024mixtralexperts}             & Mixtral-8x7B-Instruct-v0.1 & 47B  \\ \bottomrule
\end{tabular}}
\caption{LLMs used in this work.
}\label{tab:llms}
\vspace{-.3cm}
\end{table}

\begin{figure*}[!h]
\centering
    \includegraphics[width=\linewidth]{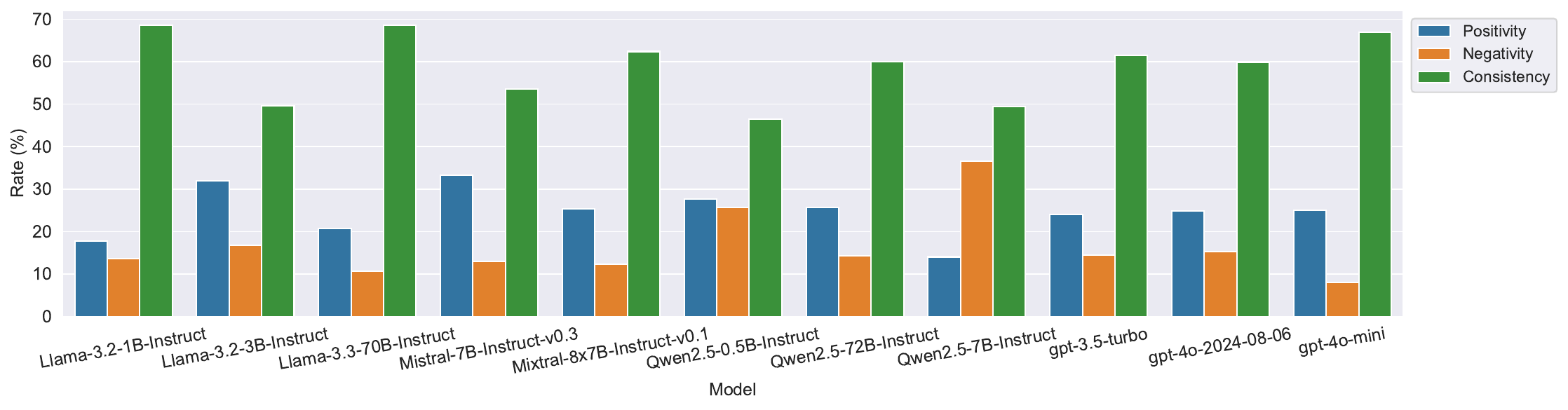}
    \vspace{-.3cm}
    \caption{Consistency, positivity, and negativity rates of LLMs' judgments on convincingness, averaged across prompts and instances in all datasets.}\label{fig:llm_dis}
    \vspace{-.3cm}
\end{figure*}

\ycr{
\subsection{Robustness of Our Findings}\label{sec:robust}
Although our annotation agreements are comparable to prior work, they remain relatively low compared to typical human annotation, as shown earlier, 
likely due to the subjective nature of the criteria. To examine the robustness of our conclusions under different conditions, we conducted an additional annotation study using a completely different setup: annotators assessed the convincingness of individual arguments independently on a Likert scale of 1-5, without pairwise comparisons. We randomly selected two batches (10 instances each) from four datasets --- \hansard{}, \deuparl{}, \dagstuhl{}, and \lynn{} --- resulting in 120 test cases (4 datasets × 10 instances × 3 test cases per instance). Each argument was rated by five crowdworkers from Prolific, and we used the average rating as the final convincingness score.}

\ycr{
The results reinforce our earlier findings: (1) Emotional content often enhances, rather than degrades, argument convincingness (positive:negative = 68:52); and (2) in over half of the cases, emotions do not substantially influence convincingness. Because it is uncommon for two arguments to receive identical average scores, we define a threshold to determine when a difference in convincingness scores between two related arguments is meaningful. With a threshold of 1 (the full scale interval), 
the consistency rate is 81.6\%; with a threshold of 0.5 (the midpoint for rounding), it drops to 57.5\% --- closely aligning with the main results reported in the paper. 
Overall, this supplemental evaluation supports the robustness of our main conclusions.
}

\section{%Human vs.\ LLMs 
\ycr{Do LLMs Behave Like Humans?}
}\label{sec:exp}

\paragraph{Models}
We select a range of recent LLMs, including both open-source and commercial models, with varying model sizes from 0.5B to 72B parameters. We experiment with \textbf{11 LLMs from 4 model families}, as detailed in Table \ref{tab:llms}. For OpenAI models, we utilize the official API,\footnote{\url{https://platform.openai.com/}} while for open-source models, we retrieve checkpoints from HuggingFace.\footnote{\url{https://huggingface.co/}} 
For all models, we set the temperature to 0.6 and the top-p value to 0.9, to ensure diverse outputs that still remain contextually relevant and logical, running each model five times. For 70B/72B models, we use 4-bit quantization. We run the models on 1 to 8 A40 GPUs, each with 48GB of memory.

\paragraph{Prompts}
We use \textbf{three prompt templates} to prompt LLMs to compare perceived 
convincingness, mirroring human instructions. 
The final judgment is determined by a majority vote from the five runs; if none is reached, the arguments are considered equally convincing. 
\textbf{Zero-shot prompts} are employed to minimize biasing effects on model responses \citep{paech2024eqbenchemotionalintelligencebenchmark} and thus better capture the models' intrinsic behavior.
As shown in Table \ref{tab:promptc} (Appendix \ref{app:prompt}), \emph{Prompt 1} instructs models to provide a label without explanation. 
\emph{Prompt 2 and Prompt 3} additionally require an explanation and include an example answer to specify the response format. To examine potential biases from examples, they feature opposite label choices and differ in perspective, with Prompt 2 favoring an objective approach and Prompt 3 adopting a more subjective and emotional stance. 

%\vspace{-.3cm}
\paragraph{LLMs exhibit a similar sensitivity to emotions when judging argument convincingness.}

Figure \ref{fig:llm_dis} presents the consistency, positivity, and negativity rates of LLMs' convincingness judgments, averaged across prompts and instances in all datasets. Like humans, \emph{LLMs show a strong tendency toward consistency}, with rates consistently exceeding positivity and negativity ($\sim$48\%-68\% vs. $\sim$10\%-36\%); all models except Qwen2.5-0.5B, Llama-3.2-3B, and Qwen2.5-7B achieve a consistency rate above 50\%;
Moreover, 
\emph{emotions more often enhance rather than degrade convincingness}, except for Qwen2.5-7B, aligning with human patterns. As shown in Figure \ref{fig:dis_prompt} (Appendix \ref{app:llm}), most models exhibit comparable rates across different prompts, except for the smallest model, Qwen2.5-0.5B, and gpt-3.5-turbo, where negativity surpasses positivity with Prompt 2. In Prompt 2's example, logical fallacy is mentioned in the explanation, which may (mis)lead models to interpret emotions as a logical fallacy.

\paragraph{However, they do not align well with humans on individual judgements.}
Table \ref{tab:llm_ranking} in Appendix \ref{app:llm} displays macro F1 scores and model rankings for LLMs in predicting argument pair convincingness rankings (column `Static’) and the resulting categories of emotional effect (column `Dynamic’) in English and German. 
The best prompt result of each model is reported to demonstrate its potential. Human and LLM labels are determined by majority votes from different annotators and runs, respectively.
Overall, all scores remain low ($\sim$0.32–0.49), indicating performance ranging from random to slightly above random in a three-way classification task. GPT4o consistently ranks first in three of four tasks, except for dynamic label prediction in English, where it ranks second.
Larger models generally align better with humans, often achieving higher F1 scores than their smaller counterparts, with the largest models (GPT4o, Llama-3.3-70B, Qwen2.5-72B) frequently ranking among the top.

%% file: structure/6_conclusion.tex
\section{Conclusion}

In this work, we examined how emotional intensity influences perceived convincingness. Using GPT4o to rephrase arguments with varying emotional impact, we developed a dynamic framework inspired by manipulation checks in psychology and social sciences. Our results show that GPT4o reliably generates counterpart arguments, preserving meaning while altering emotional tone.
For both humans and LLMs, convincingness is largely unaffected by emotions. However, when emotions do play a role, they more often enhance rather than weaken convincingness, particularly in political debates, where emotional appeal is frequently used as a persuasive strategy. Additionally, while LLMs broadly mirror human patterns, they struggle to capture emotional nuances.

Future research could explore \textbf{when and how} emotions influence convincingness across argument types. Investigating \textbf{specific emotions} \citep{greschner2024fearfulfalconsangryllamas} or \textbf{justified vs. unjustified emotions} and their persuasive effects may provide deeper insights. Enhancing LLMs' ability to capture emotional nuances through improved prompts or fine-tuning could further strengthen their reliability in evaluating emotional arguments.

\section*{Limitations \& Ethical concerns}
While our study provides insights into the relationship between emotional intensity and argument convincingness, several limitations should be acknowledged:
(1) We rely on a single model, GPT4o, for synthetic argument generation. While GPT4o demonstrates strong capabilities in controlled text modification, exploring multiple models could provide a more comprehensive understanding of how different architectures handle emotional rephrasing.
(2) We focus only on two languages, English and German. Expanding to additional languages, particularly those with different rhetorical traditions or cultural perspectives on emotional persuasion, would offer a broader cross-linguistic perspective.
(3) The topics of arguments differ across text domains, which may introduce variability in how emotional intensity interacts with convincingness. Ensuring more comparable topics across domains would help isolate the individual effects of topic and text domain, leading to a more precise analysis.
\ycr{(4) The dataset is relatively small, and the annotation agreement is low, which may limit the generalizability of our findings. However, with an additional annotation study (§\ref{sec:robust}), we were able to replicate the main observations.}
(5) We do not distinguish between different types of emotions (e.g., anger, joy, fear) or between justified and unjustified emotions, both of which could have varying impacts on argument convincingness. Future work could explore how different kinds of emotions influence persuasion to gain a more nuanced understanding of their effects. (6) We experiment with only three prompts to evaluate model responses, which may not fully reflect LLM performance. A broader range of prompts could yield more stable results.

A potential ethical concern arises from the possibility of leveraging the dataset to develop politically motivated agendas that rely on emotional appeal rather than factual reasoning. Since emotions can influence perceived convincingness, there is a risk that political actors or interest groups may use this dataset to craft emotionally charged arguments that manipulate public opinion rather than inform it. This could contribute to misinformation, polarization, and biased discourse, particularly in sensitive political debates.

We used ChatGPT solely for text refinement while writing this paper. All annotators provided consent for research use of their annotations via Google Forms.

%% file: structure/999_appendix.tex
\section{Pre-annotation and classifiers}\label{app:pre}
\paragraph{Pre-annotation}
We start with the \textbf{Second Reading debates of Bills},\footnote{\url{https://www.parliament.uk/about/how/laws/passage-bill/commons/coms-commons-second-reading/}} where the members debate the main principles of a certain Bill. The advantages of using such debates are: (i) the stance of an argument can be easily identified based on whether they support %for 
the Bills; (ii) debates can be paired with brief Bill introductions,\footnote{e.g., the `long title' on page \url{https://bills.parliament.uk/bills/3858}} providing clear argument topics; and (iii) the arguments 
focus on Bill principles, with fewer discussions on specific amendments and clauses, which require less contextual awareness than other Bill debates like the ones for the Committee Stage.\footnote{\url{https://www.parliament.uk/about/how/laws/passage-bill/commons/coms-commons-comittee-stage/}}
We choose five Bills, including topics relevant to animal welfare and parental leave (see Table \ref{tab:bill} for the Bill introductions), 
which may be easier to annotate and more likely to have emotional arguments.

Three annotators label 245 texts from these debates for \textbf{three layers}: (\emph{L1}) 
whether the text evokes emotions, (\emph{L2}) whether the text contains standalone arguments, and (\emph{L3}) the stance of the text toward the Bill. \emph{L1} and \emph{L2} are labeled `0' (for answer `no') or `1' (for `yes'). If \emph{L2} is labeled `1', annotators proceed to label \emph{L3}, which has four options: `0' for support, `1' for opposition, `2' for inability to identify stance without additional context, and `3' for a neutral stance suggesting additional amendments or policies. Besides, 40 texts from the pilot annotation are also annotated for \emph{L1} and \emph{L2}.  
To potentially speed up the annotation process, the 285 texts are selected from those judged as both emotional and argumentative by GPT4o. Here, we prompt GPT4o with simple questions such as \emph{Does this text try to convince readers something?} and \emph{Is this text emotional?'}.

40 of the outputs are jointly labeled by all annotators, achieving average Cohen's Kappa of 0.622 for \emph{L1}, 0.674 for \emph{L2}, and 0.762 for \emph{L3} across annotator pairs. 
As shown in the `Question' column of Table \ref{tab:pre}, GPT4o already achieves a high precision of 0.82 in detecting argumentative texts using simple prompts. However, its precision for emotional text classification is still low (0.53).

We then convert the annotations for \emph{L3} to \emph{L3$^{*}$}, where we pair argument pairs based on their topics and stances. The categories include: `different topic' for pairs with different topics (from different Bills), `different stance' for pairs with the same topic but different stances, and `same' for pairs with the same topic and stance.

The number of texts annotated for each layer and the corresponding label distribution\se{s} are summarized in Table \ref{tab:pre} (left). 

\paragraph{Automatic Pipeline}
We develop three classifiers based on GPT4o 
to automatically identify the argument pairs needed. The pipeline is as follows: 
\begin{enumerate}[]
    \item \textbf{Argumentative text classification}: our goal is to have a \textbf{high precision} classifier since we have sufficient candidate texts. We find that when we ask GPT-4o to provide the major claim, evidence, and reasoning connecting the evidence to the major claim in the text, its precision increases from 0.82 to 0.96, as shown in the `Argumentative' row of Table \ref{tab:pre}. 
    
    We then retain texts judged as argumentative for \hansard{} using this prompt, while for \deuparl{}, we use a German translation of the same prompt. The overall performance of GPT4o on German data is assessed after completing the stance agreement classification task (see below).

    \item \textbf{Stance agreement classification}: 
    To enable the flexible selection of classifiers with specific performance characteristics (e.g., high recall, high precision), we introduce a parameter into the prompt, with its threshold optimized to achieve different specialized performance levels.
    To do so, we ask GPT4o to rate the likelihood that two given arguments address the same topic and share the same stance on a Likert scale from 0 to 100. We randomly sample 600 argument pairs (with a 2:1:1 ratio for the three categories of \emph{L3$^{*}$}) from the dataset, ‘optimize’ the threshold of ratings for the `same’ category 
    using argument pairs from two Bills, and test the performance on the remaining three Bills to prevent data leakage. We evaluate all possible combinations of Bills for the training and test sets.
    We observe that as the threshold increases, precision on the `same’ category ($P_{same}$) consistently improves, while macro F1 begins to decrease beyond certain thresholds. With a threshold of 100, $P_{same}$ reaches 0.92, but F1 is very low at 0.45. Therefore, we select a threshold of 90 as a more balanced trade-off, achieving $P_{same}=0.81$ and $\textit{F1}=0.76$, to obtain more candidates that are still highly likely to be true positives. 
    
    For \hansard{}, we retain the argument pairs labeled as belonging to the `same' category using this threshold. For \deuparl{}, we apply the German translation of the prompt with the same threshold to identify argument pairs. One annotator evaluates 50 candidates from the outputs of steps 1 and 2: no argument is labeled as non-argumentative, while 12 argument pairs are identified as false positives in the stance agreement task, yielding $P_{same}=0.76$. This value is only 4 percent points lower than the result on English data. Consequently, we retain these prompt settings for the German data.
    
    \item  \textbf{Emotional text classification}: we aim for a \textbf{balanced} classifier because we also need non-emotional arguments. Since this is a subjective task, we ask GPT4o to rate how likely it can feel the emotions 
    in the texts on a \se{L}ikert scale of 0-100, and then `optimize' the threshold of the rates for the `emotional' category on 70\% of the data and check how it performs on the remaining 30\%. Overall, with this step, we can improve the macro F1 to 0.74-0.81 (averaged over three rounds of data splitting), depending on the gold from different annotators. The best threshold for two annotators is 75, while that for the other is 85, so we use the threshold 75 to represent the majority, which has a macro F1 of 0.75, averaged across the three annotators. 

    We use this threshold to select the argument pairs for \hansard{}. For \deuparl{}, we further optimize the threshold using a small-scale set of human annotations and adjust it to 85. This setting is then used to label the binary emotions of arguments. 

\end{enumerate}

\begin{table}[!ht]
\resizebox{\linewidth}{!}{%
\begin{tabular}{@{}lcccc@{}}
\toprule
                               & \multicolumn{2}{c}{Pre-Annotation} & \multicolumn{2}{c}{Automatic Pipeline}   
                               \\
                               & \#                 & \%  & Question  & `Optimized' \\ \midrule
\multicolumn{3}{l}{\emph{L1 - emotion} }                                             \\ \midrule
Emotional                      & 151                & 53.0 & 0.53 (P)  & \multirow{2}{*}{0.75 (F1)} \\ 
Non-emotional                  & 134                & 47.0 & -  \\ \bottomrule
\multicolumn{3}{l}{\emph{L2 - argument}}                                         \\ \midrule
Argumentative                  & 234                & 82.1 & 0.82 (P) & 0.96 (P)  \\
Non-argumentative              & 51                 & 17.9 & - & - \\ \bottomrule
\multicolumn{3}{l}{\emph{L3 - stance} }                                           \\ \midrule    
Support                        & 170                & 72.6 & - \\
Opposition                     & 2                  & 0.9 & -  \\
Neutral                        & 29                 & 12.4  & -\\
Irrelevant                     & 16                 & 14.1& - \\ \midrule
\multicolumn{3}{l}{\emph{L3$^{*}$ - pair stance}}                                            \\ \midrule   
Same           & 2,905            & 8.9  & -   & \multirow{3}{*}{\makecell{0.80 ($P_{same}$) \\ 0.75 (F1)}} \\
Different stance & 3,325              & 10.2 & - \\
Different topic                & 26,486             & 81.0 & - \\ \midrule
Total                          & 32,716             & 100  & - \\ \bottomrule
\end{tabular}}
\caption{Number of texts annotated for each layer and category (\#) and the corresponding label distribution (\%). Performance of GPT4o on the binary emotion classification, argument identification, and stance agreement detection tasks used for automatically identifying the target argument pairs.}\label{tab:pre}
\end{table}

\begin{table*}[!ht]
\resizebox{\linewidth}{!}{
\begin{tabular}{@{}l@{}}
\toprule
\emph{Introduction}                                                                                                                                                                         \\ \midrule
A Bill to Prohibit the export of certain livestock from Great Britain for slaughter.                                                                                                 \\ \midrule
\makecell[l]{A Bill to create offences of dog abduction and cat abduction and to confer a power to make corresponding provision  \\ relating to the abduction of other animals commonly kept as pets.} \\ \midrule
A Bill to make provision about leave and pay for employees with responsibility for children receiving neonatal care.                                                                   \\ \midrule
A Bill to prohibit the import and export of shark fins and to make provision relating to the removal of fins from sharks.                                                            \\ \midrule
A Bill to prohibit the sale and advertising of activities abroad which involve low standards of welfare for animals.                                                                 \\ \bottomrule
\end{tabular}}
\caption{The introductions of the five Bills selected in \protect\bill{}.}\label{tab:bill}
\end{table*}

% \usepackage{booktabs}
\begin{table*}[]
\resizebox{\linewidth}{!}{
\begin{tabular}{@{}l|l@{}}
\toprule
English                                                                                                                                                                                  & German                                                                                                                                                                                     \\ \midrule
\makecell[l]{iran, integrat, ukraine, russia, asylum,\\ deportation, israel, gaza, expulsion, \\ displacement, migration, migrant, \\immigrant, refugee, palestine,invasion,\\ repatriation, hamas, hisbollah} & \makecell[l]{ukraine, russland, migrant, \\ immigrant, flüchtling, asyl,\\ gaza, iran, palästina, \\israel, krieg, invasion, \\sanktionen, waffenlieferungen, friedensverhandlungen, \\kriegsverbrechen, flüchtlingskrise, nato,\\ energieversorgung, vertreibung, migrationspolitik,\\ asylverfahren, grenzsicherung, integration, \\abschiebung, aufenthaltsgenehmigung, menschenhandel, \\seenotrettung, rückführung, schutzstatus, \\waffenstillstand, raketenangriffe, besatzung, \\zwei-staaten-lösung, friedensprozess, intifada, \\ hamas, hisbollah, menschenrechte, un-resolution
} \\ \bottomrule
\end{tabular}
}
\caption{Keywords used to filter debates for \hansard{} and \deuparl{}.}\label{tab:keywords}
\end{table*}

% \usepackage{booktabs}
\begin{table*}[]
\centering
%\resizebox{!}{\linewidth}{
\begin{tabularx}{\linewidth}{X}
\toprule
\emph{Remove Emotion Prompt}  \\ \midrule
====\textbf{System Prompt}=====\\ I will give you an argumentative text that **can** appeal to emotion.    \\ \\ Your task is to generate an argument with the same stance for the same topic **without emotional language**, by rephrasing the text but maintaining a similar style and length. \\ \\ Briefly explain why the rewritten argument no longer evokes emotions.\\ \\ Answer in the following way:\\ Generated argument: \\ Explanation:\\ ====\textbf{User Prompt}=====\\ Text: \{original argument\}  \\ \midrule
\emph{Add Emotion Prompt}  \\ \midrule
====\textbf{System Prompt}=====\\ I will give you an argumentative text that **cannot** appeal to emotion.\\     \\ Your task is to generate an argument with the same stance on the same topic **with emotions**, by rephrasing the text but maintaining a similar style and length. \\ \\ Briefly explain why the rewritten argument can evoke emotions now.\\ \\ Answer in the following way:\\ Generated argument: \\ Explanation:\\ ====\textbf{User Prompt}=====\\ Text: \{original argument\}                \\ \bottomrule
\end{tabularx}
%}
\caption{Prompts used to remove/add emotions for synthetic arguments.}\label{tab:prompt_synthetic}
\end{table*}

\section{Arguments from others}\label{app:other}
\paragraph{\dagstuhl{}} 
\citet{wachsmuth-etal-2017-computational} collected human ratings on a Likert scale of 1–3 for multiple dimensions of argument quality, including argument effectiveness (convincingness)\footnote{“Argumentation is effective if it persuades the target audience of (or corroborates agreement with) the author’s stance on the issue.” — \citet{wachsmuth-etal-2017-computational}} and emotional appeal. These ratings were applied to 304 argumentative texts from \citet{habernal-gurevych-2016-argument}, which were sourced from a textual debate portal in \textbf{English}. We retain only those arguments whose average convincingness rating (across the three annotators) exceeds 1.5. 
Next, we pair arguments that share the same stance on the same topics and calculate the absolute differences in their emotional appeal ratings. From these pairs, we randomly select 10 topics and then retain the 5 argument pairs with the largest absolute differences in emotional appeal for each topic.

\paragraph{\lynn{}}
\citet{greschner2024fearfulfalconsangryllamas} collected
discrete emotion labels from a reader respective (e.g. joy, disgust etc.) for 300 \textbf{German} arguments associated with 30 statements, drawn from \citet{velutharambath_wuehrl_klinger_2024a}. Each argument was annotated by three annotators. We interpret the number of annotations marking the argument as containing specific emotions (rather than `no emotion') as its emotion score. E.g., if three annotators identify specific emotions in the argument, its emotion score would be 3. Using a procedure similar to the one employed for \dagstuhl{}, we pair arguments referencing the same statement, randomly select 25 statements, and then retain the 
two argument pairs per statement that exhibit the greatest differences in emotion scores.

\begin{table*}[]
\centering
\resizebox{.6\linewidth}{!}{
\begin{tabular}{@{}llllll@{}}
\toprule
     & \dagstuhl{} & \bill{} & \hansard{} & \lynn{} & \deuparl{} \\ \midrule
\multicolumn{6}{l}{\emph{Increase}}                               \\ \midrule
\rz{}   & \textbf{-0.06}   & \textbf{0.15}         & \textbf{0.05}   & \textbf{-0.38}  & \textbf{0.32}   \\
\rth{} & -0.18    & -0.21        & -0.31  & -0.46   & -0.38  \\ \midrule
\multicolumn{6}{l}{\emph{Decrease}    }                        \\ \midrule
\ro{} & -0.12 & -0.21 & -0.03 & 0.08 & -0.19    \\
\rtw{} & \textbf{0.36}    & \textbf{0.27}         & \textbf{0.29}   & \textbf{0.76}   & \textbf{0.25} \\ \bottomrule
\end{tabular}}
\caption{BWS scores for the 4 argument groups: \rz{}, \ro{}, \rtw{} and \rth{}, derived from the majority votes of the annotation for pairwise comparisons of emotional intensity. `Increase'/`Decrease' denotes the direction to increase/decrease the perceived emotional intensity.}\label{tab:bws}
\vspace{-.3cm}
\end{table*}

\section{Prompts}\label{app:prompt}
Table \ref{tab:prompt_synthetic} presents the prompts used to introduce/remove emotions. Table \ref{tab:promptc} illustrates the prompts used for evaluating argument convincingness.

% \usepackage{booktabs}
\begin{table*}
\footnotesize
\resizebox{\linewidth}{!}{
\begin{tabular}{@{}cl@{}}
\toprule
\multicolumn{2}{l}{Prompt Template}                                                                                                                                                                                                                                                                                                                                                                                                                                                                                                                                                                                                                                                                                     \\ \midrule
Shared & \begin{tabular}[c]{@{}l@{}}Below, you will find one pair of argumentative texts discussing the same topic with the same stance. The topic may be a binary \\choice, a bill from UK parliamentary debates, or a simple statement. Both arguments either support or oppose the topic, or they \\favor one side if the topic involves a binary choice.\\ \\ Your task is to evaluate each pair to determine **which argumentative text you find more convincing**. There are three label options:\\ 0 (Both arguments are equally convincing.)\\ 1 (Argument 1 is more convincing.)\\ 2 (Argument 2 is more convincing.)\\ \\ **Note**: Truncated sentences or grammatical errors should be **ignored**.\end{tabular} \\ \midrule
1      & \begin{tabular}[c]{@{}l@{}}Please answer your label option **without** any explanations.\\ \\ \{text\}\end{tabular}                                                                                                                                                                                                                                                                                                                                                                                                                                                                                                                                                                                            \\ \midrule
2      & \begin{tabular}[c]{@{}l@{}}Please answer your label option and briefly explain why you choose this label.\\ \\ \{text\}\\ \\ Below is an example answer for you; please follow this format in your response.\\ Label: 2\\ Explanation: because Argument 2 provides more statistics supporting the claim, while Argument 1 contains logical fallacies.\end{tabular}                                                                                                                                                                                                                                                                                                                                             \\ \midrule
3      & \begin{tabular}[c]{@{}l@{}}Please answer your label option and briefly explain why you choose this label.\\ \\ \{text\}\\ \\Below is an example answer for you; please follow this format in your response.\\ Label: 1\\ Explanation: Argument 1 is more convincing, because I totally agree with its point and it evokes my empathy.\end{tabular}                                                                                                                                                                                                                                                                                                                                                                                                                                                            \\ \bottomrule
\end{tabular}}
\caption{Prompt templates for comparing the convincingness of an argument pair. The {text} field contains the two arguments and their topic. The complete prompt is formed by combining the text in the `Shared' row with the text in the corresponding indexed row. For example, Prompt 1 consists of the text from both the `Shared' row and row `1'.}\label{tab:promptc}
\end{table*}

\section{Annotation Interface}\label{app:anno}
\begin{figure*}
    \includegraphics[width=\linewidth]{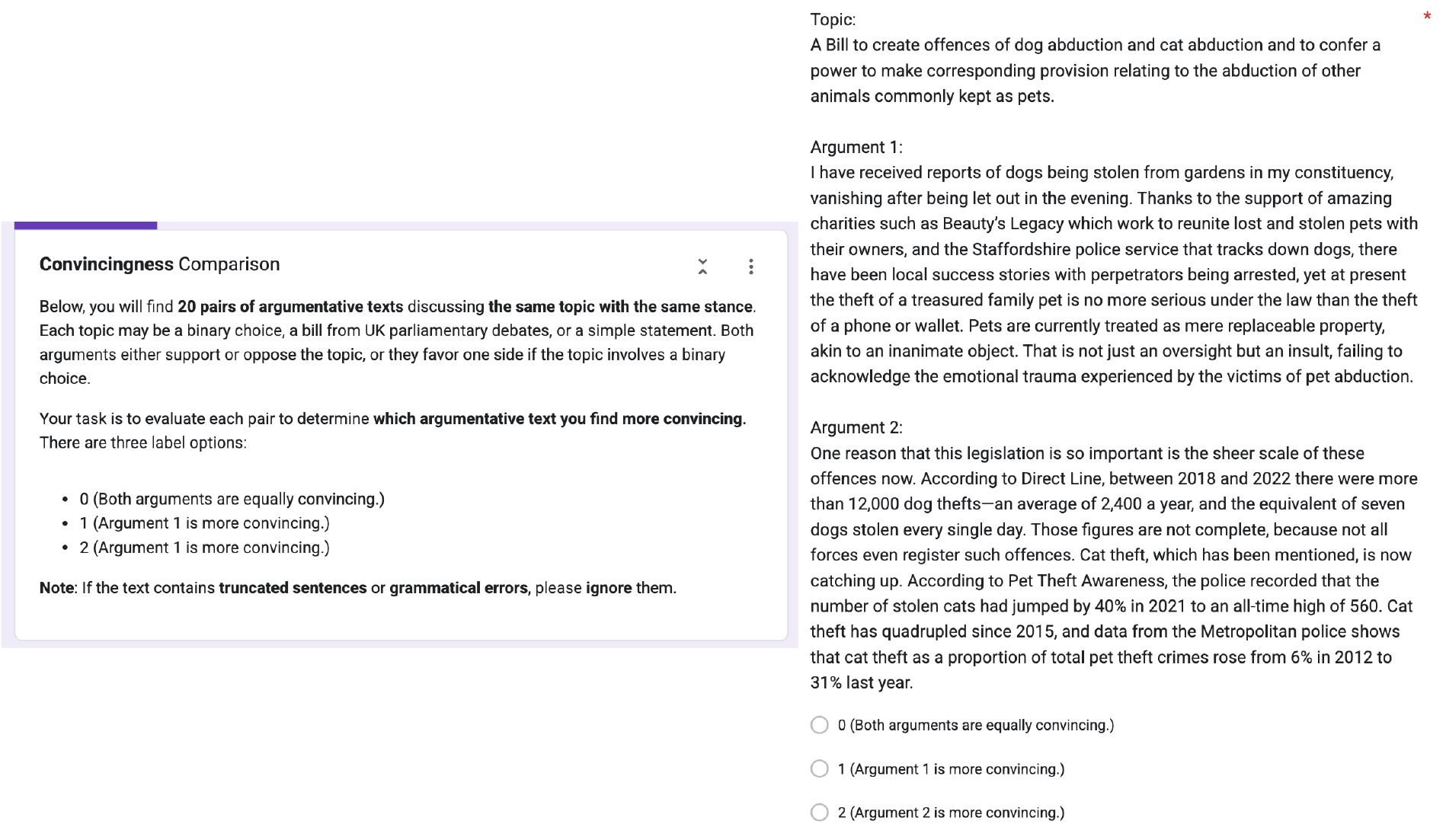}
    \includegraphics[width=\linewidth]{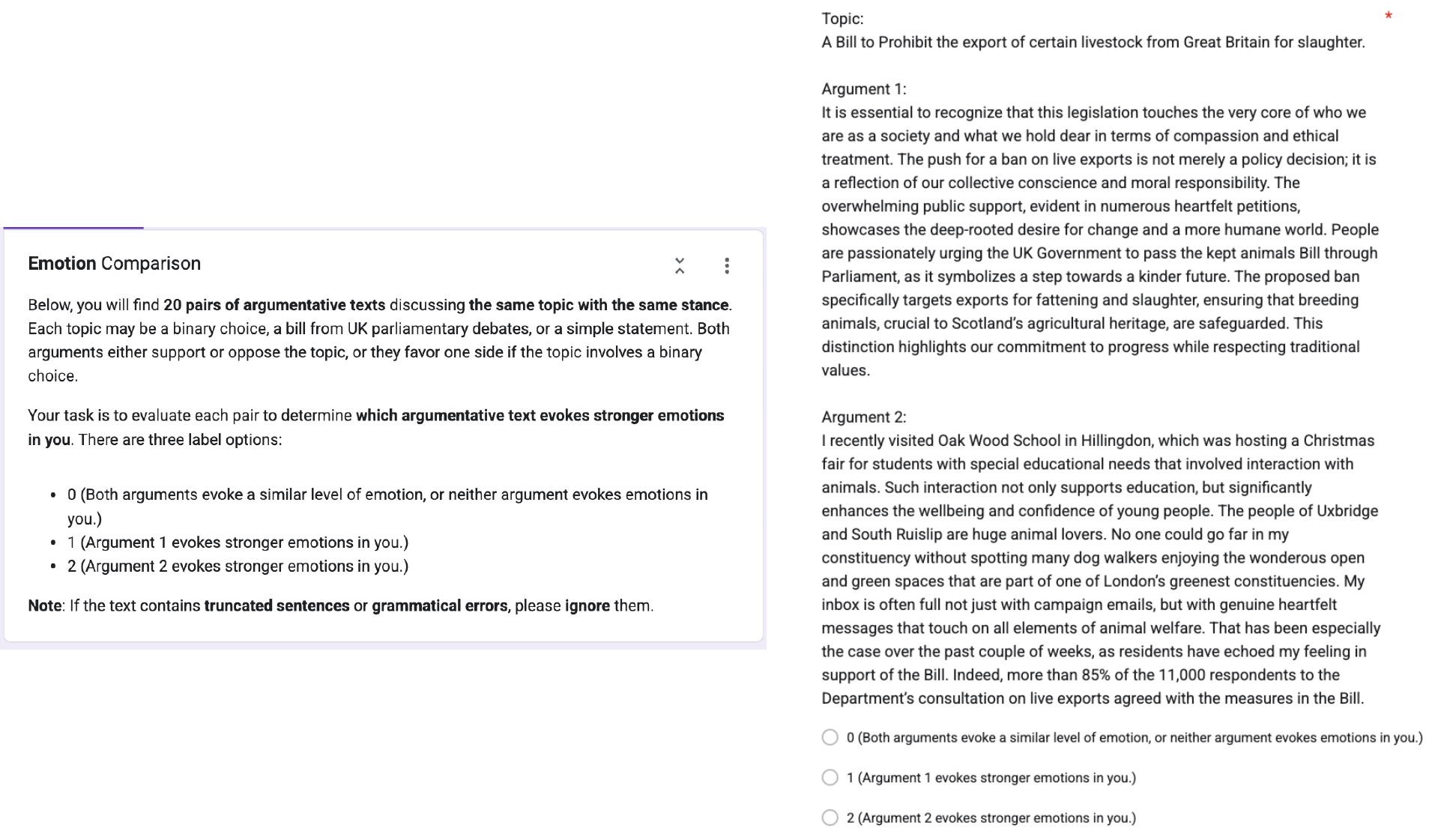}
    \caption{Screenshots of the annotation interface for convincingness (top) and emotion (bottom) comparison.}\label{fig:anno}
\end{figure*}

Figure \ref{fig:anno} shows the screenshots of the annotation interface for convincingness (top) and emotion (bottom) comparisons. We collect the annotations via Google Forms\footnote{\url{https://docs.google.com/forms/}} for crowdsourcing annotators.

\section{Examples}\label{app:exa}
Table \ref{tab:pos} and \ref{tab:neg} provide example instances from \hansard{} and \lynn{}, where emotions have a positive and negative impact, respectively. 

\begin{table*}[!ht]
\centering
\begin{tabularx}{\textwidth}{ X | X }
\toprule
\multicolumn{2}{l}{\textbf{Topic}: The public supports the UK's aid for Ukrainian refugees} \\ \midrule
\rz{}  & \ro{}  \\ \midrule
Members across this House are determined that we, as a country, should open our arms to these people, and this determination has been on full display today. The scenes of devastation and human misery inflicted by President Putin’s barbarous assault on what he calls “Russia’s cousins” in Ukraine have unleashed a tidal wave of solidarity and generosity across the country. British people always step forward and step up in these moments, and since the first tanks rolled into Ukraine, they have come forward in droves with offers of help: community centres have been flooded with critical supplies; the Association of Ukrainians in Great Britain has received millions in donations; and charities such as the Red Cross have been overwhelmed with people giving whatever they can. The outpouring of public support has been nothing short of remarkable. & While this Government, and this whole House, have risen to the occasion with our offer of support to Ukrainians fleeing war, our lethal aid and our stranglehold on economic sanctions on Russia have clearly shown that we will keep upping the ante to ensure that Putin fails. As Members have argued today, it has been abundantly clear in recent days that we can and must do more. It is exactly right, therefore, that my right hon. Friend the Secretary of State for Levelling Up, Housing and Communities set out on Monday the new and uncapped sponsorship scheme, Homes for Ukraine. It is a scheme to allow Ukrainians with no family ties to the UK to be sponsored by individuals or organisations that can offer them a home. It is a scheme that draws not only on the exceptional good will and generosity of the British people, but one that gives them the opportunity to help make a difference.                                                                                                                                                        \\ \midrule
\rth{}   & \rtw{} \\ \midrule
Members of this House have expressed a commitment to welcoming individuals from Ukraine. The recent conflict initiated by President Putin has resulted in significant destruction in Ukraine, prompting a substantial response of support across the country. British citizens have actively contributed since the conflict began, with community centers collecting essential supplies, the Association of Ukrainians in Great Britain receiving financial contributions, and charities like the Red Cross witnessing increased donations.  & In these trying times, the Government and this entire House have demonstrated unwavering courage and compassion by extending our support to Ukrainians escaping the horrors of war. Our determined provision of lethal aid and the relentless imposition of economic sanctions on Russia are powerful affirmations that we will stop at nothing to ensure Putin's defeat. As Members have passionately discussed today, the urgency to do even more has never been clearer. That is why it is so heartening that my right hon. Friend the Secretary of State for Levelling Up, Housing and Communities announced on Monday the new and limitless Homes for Ukraine sponsorship scheme. This initiative opens its arms to Ukrainians without family connections in the UK, allowing them to be warmly embraced by individuals or organizations ready to offer them a sanctuary. It is a testament not only to the extraordinary kindness and generosity of the British people but also to their deep desire to make a meaningful impact in the lives of those in desperate need. \\ \bottomrule
\end{tabularx}
\caption{An example instance from \hansard{} where emotions have a \textbf{positive} impact on argument convincingness.
}\label{tab:pos}
\end{table*}

\begin{table*}[!ht]
\centering
\begin{tabularx}{\textwidth}{ X | X }
\toprule
\multicolumn{2}{l}{Topic: Haie können Krebs bekommen.}    \\ \midrule
\rz{}  & \ro{}  \\  \midrule
Haie sind mehrzellige Lebewesen, wie auch der Mensch. Die Beonderheit von mehrzelligen Lebewesen ist, dass die Zellen sich sowohl stark spezialisieren und untereinander vernetz kommunizieren. Damit werden sie anfällig für bestimmte Zelldefekte, die sich über die genannte Struktur fortpflanzen und den Krebs ausmachen. Haie verfügen, wie auch der Mensch und überhaupt alle mehrzelligen Lebewesen, über nur eine sehr eingeschränkte Möglichkeit diese Defekte zu korrigieren und aufzuhalten, damit können beide gleichermaßen Krebs bekommen & Da auch Fische Krebs bekommen können, ist es auch möglich, dass Haie Krebs bekommen können. Dieser wird durch mutierte Zellen ausgelöst, weshalb dies auch bei Fischarten ausgelöst werden kann. Krebs ist eine weit verbreitete und häufige Krankheit, weshalb Krebs durch Wissenschaftler auch bereits bei Haien festgestellt werden konnte.\\ Krebs kann außerdem auch durch verschiedene Umweltfaktoren wie Umweltverschmutzung ausgelöst werden, diesem Risiko sind Haie ja durchaus ausgesetzt. Deshalb ist die Gefahr einer Erkrankung auch nicht gerade gering.  \\ \midrule
\rth{}  & \rtw{}  \\ \midrule
Haie, ebenso wie Menschen, sind mehrzellige Organismen. Eine charakteristische Eigenschaft solcher Organismen ist die Spezialisierung und Vernetzung ihrer Zellen. Diese Struktur macht sie anfällig für Zellfehler, die sich ausbreiten und zu Krebs führen können. Haie und Menschen besitzen nur begrenzte Mechanismen zur Korrektur und Kontrolle dieser Defekte, was bedeutet, dass beide Arten gleichermaßen anfällig für Krebs sind.  & Die Vorstellung, dass Haie - diese majestätischen und oft missverstandenen Kreaturen der Meere - an Krebs erkranken können, ist zutiefst beunruhigend. Diese Krankheit, die durch die heimtückische Mutation von Zellen verursacht wird, hat bereits viele Fischarten heimgesucht. Die Tatsache, dass auch Haie, die Könige der Ozeane, nicht sicher vor dieser grausamen Krankheit sind, ist erschütternd. Angesichts der weit verbreiteten Umweltverschmutzung, die unsere Ozeane verschlingt, sind Haie einem erheblichen Risiko ausgesetzt, an Krebs zu erkranken. Es ist traurig und alarmierend, dass diese beeindruckenden Tiere, die seit Millionen von Jahren die Meere durchstreifen, nun durch menschliche Einflüsse bedroht sind.
\\ \bottomrule
\end{tabularx}
\caption{An example instance from \lynn{} where emotions have a \textbf{negative} impact on argument convincingness.
}\label{tab:neg}
\end{table*}

\section{LLM}\label{app:llm}
\begin{figure*}[]
    \centering
    \includegraphics[width=\linewidth]{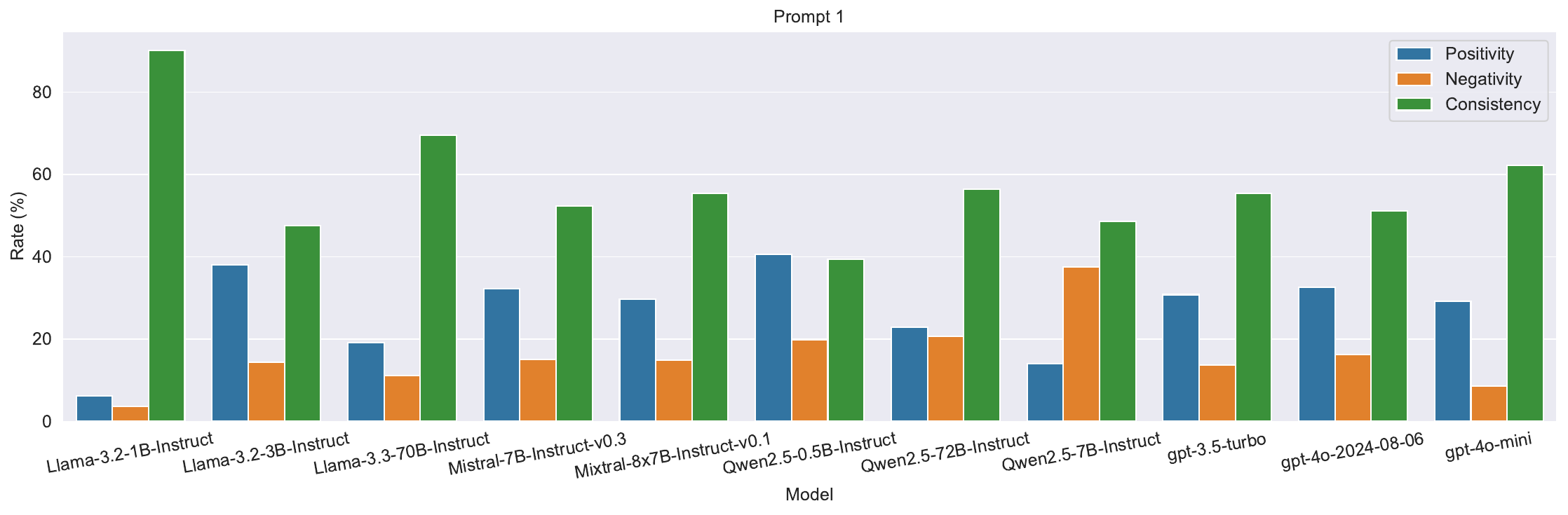}
    \includegraphics[width=\linewidth]{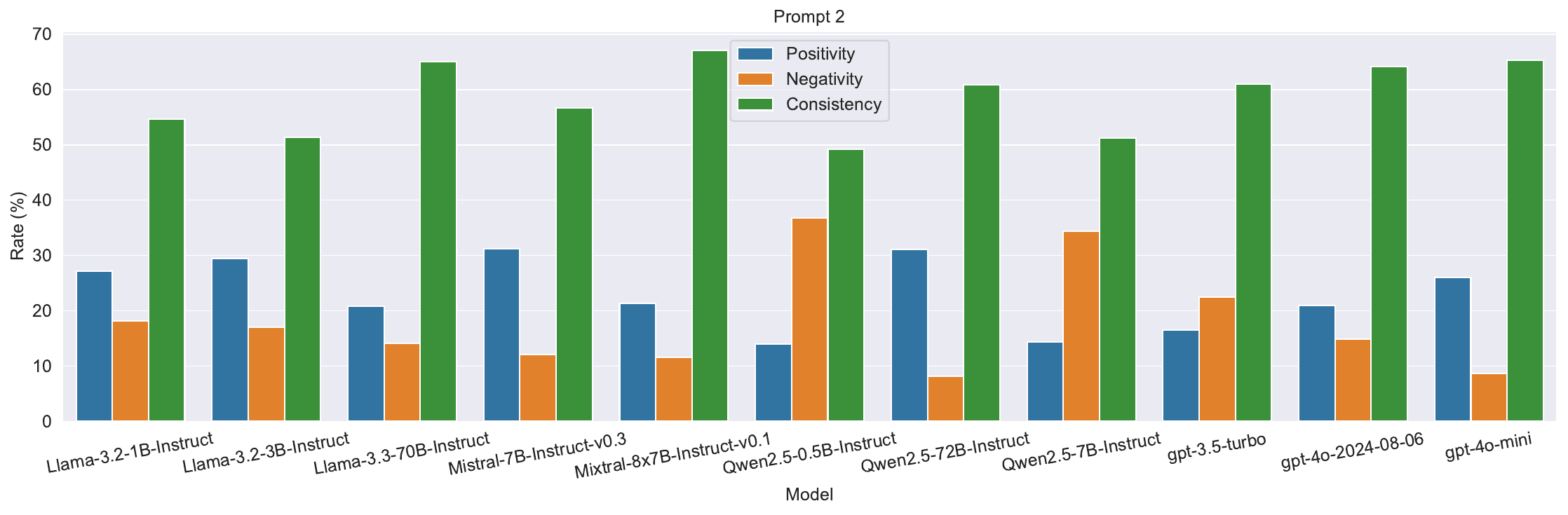}
    \includegraphics[width=\linewidth]{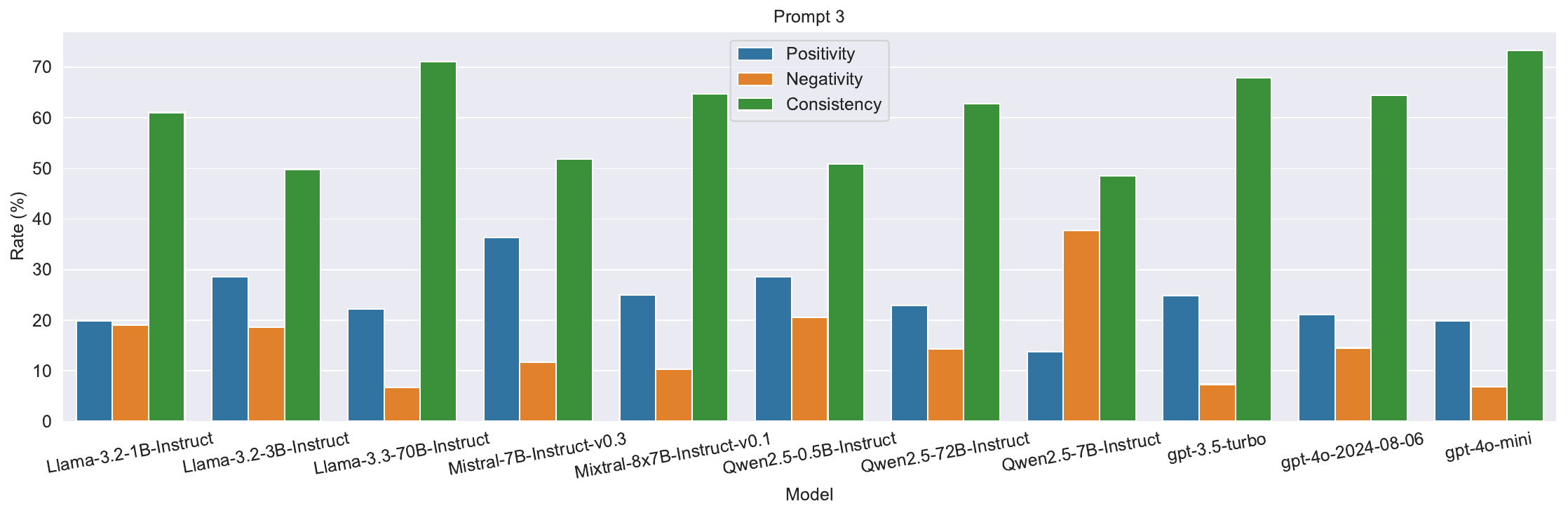}
    \caption{Consistency, positivity and negativity rates of LLMs with different prompts, averaged across instances in all datasets.}\label{fig:dis_prompt}
\end{figure*}

Figure \ref{fig:dis_prompt} illustrates the consistency, positivity and negativity rates of LLMs with different prompts, averaged across instances in all datasets. Table \ref{tab:llm_ranking} displays macro F1 scores and model rankings for LLMs in predicting convincingness rankings of argument pairs ('Static') and the resulting categories of emotional effect (`Dynamic') in English and German.

\begin{table*}[]
\resizebox{\linewidth}{!}{
\begin{tabular}{lcccc|cccc}
\toprule
                           & \multicolumn{4}{c|}{\textbf{EN}}               & \multicolumn{4}{c}{\textbf{DE}}               \\ 
\textbf{Model}                      & Static & Ranking & Dynamic & Ranking & Static & Ranking & Dynamic & Ranking \\ \midrule
gpt-4o-2024-08-06          & \textbf{0.486}  & 1       & 0.411   & 2       & \textbf{0.443}  & 1       & \textbf{0.447}   & 1       \\
Llama-3.3-70B-Instruct     & 0.417  & 2       & \textbf{0.415}   & 1       & 0.372  & 2       & 0.392   & 4       \\
gpt-4o-mini                & 0.416  & 3       & 0.392   & 5       & 0.35   & 4       & 0.394   & 3       \\
Qwen2.5-72B-Instruct       & 0.398  & 4       & 0.398   & 4       & 0.357  & 3       & 0.41    & 2       \\
gpt-3.5-turbo              & 0.39   & 5       & 0.382   & 6       & 0.338  & 6       & 0.381   & 6       \\
Mixtral-8x7B-Instruct-v0.1 & 0.368  & 6       & 0.376   & 7       & 0.35   & 5       & 0.387   & 5       \\
Mistral-7B-Instruct-v0.3   & 0.367  & 7       & 0.407   & 3       & 0.288  & 8       & 0.36    & 9       \\
Llama-3.2-3B-Instruct      & 0.322  & 8       & 0.32    & 10      & 0.281  & 10      & 0.367   & 8       \\
Qwen2.5-0.5B-Instruct      & 0.308  & 9       & 0.342   & 9       & 0.284  & 9       & 0.344   & 10      \\
Qwen2.5-7B-Instruct        & 0.304  & 10      & 0.346   & 8       & 0.319  & 7       & 0.373   & 7       \\
Llama-3.2-1B-Instruct      & 0.286  & 11      & 0.309   & 11      & 0.274  & 11      & 0.343   & 11 \\ \bottomrule     
\end{tabular}}
\caption{Macro F1 scores and model rankings for LLMs in predicting convincingness rankings of argument pairs ('Static') and the resulting categories of emotional effect (`Dynamic') in English and German. For each model, we present the best prompt result to highlight its potential. Human and LLM labels are determined by majority votes from different annotators and rounds, respectively.}\label{tab:llm_ranking}
\end{table*}